%% file: main.tex
\pdfoutput=1

\documentclass[11pt]{article}

\usepackage[final]{acl}

\usepackage{times}
\usepackage{latexsym}

\usepackage[T1]{fontenc}

\usepackage[utf8]{inputenc}

\usepackage{microtype}

\usepackage{inconsolata}

\usepackage{graphicx}
\usepackage{graphics}
\usepackage{amsmath}
\usepackage{amssymb}
\usepackage{mathtools}
\usepackage{amsthm}
\usepackage{booktabs}
\usepackage{multirow}
\usepackage{multicol}
\usepackage{url}
\usepackage{subfigure}
\usepackage[bottom]{footmisc}

\newtheorem{theorem}{Theorem}[section]

\newtheorem{lemma}[theorem]{Lemma}

\newtheorem{definition}[theorem]{Definition}

\title{Mitigating the Linguistic Gap with Phonemic Representations \\ 
for Robust Cross-lingual Transfer}

\author{
 \textbf{Haeji Jung\textsuperscript{1}},
 \textbf{Changdae Oh\textsuperscript{2}},
 \textbf{Jooeon Kang\textsuperscript{3}},
 \textbf{Jimin Sohn\textsuperscript{4}},
\\
 \textbf{Kyungwoo Song\textsuperscript{5}},
 \textbf{Jinkyu Kim\textsuperscript{1}},
 \textbf{David R. Mortensen\textsuperscript{6}}
\\
\\
 \textsuperscript{1}Korea University,
 \textsuperscript{2}University of Wisconsin--Madison,
 \textsuperscript{3}Sogang University,
 \\
 \textsuperscript{4}GIST,
 \textsuperscript{5}Yonsei University,
 \textsuperscript{6}Carnegie Mellon University
\\
 \small{
   \textbf{E-mail:} \href{mailto:gpwl@korea.ac.kr}{gpwl0709@korea.ac.kr}, \href{mailto:dmortens@cs.cmu.edu}{dmortens@cs.cmu.edu}
 }
} 

\begin{document}
\maketitle

\input{sec/0_abstract}
\input{sec/1_intro}

\input{sec/2_related_works}

\input{sec/3_experiments}
\input{sec/4_numerical_analysis}
\input{sec/5_theoretical_analysis}
\input{sec/6_conclusion_limitation}

\input{sec/10_impact}
\input{sec/7_ethics}
\input{sec/8_acknowledgement}

\bibliography{main}

\appendix
\input{sec/9_appendix}

\end{document}

%% file: sec/0_abstract.tex
\begin{abstract}

Approaches to improving multilingual language understanding often struggle with significant performance gaps between high-resource and low-resource languages. While there are efforts to align the languages in a single latent space to mitigate such gaps, how different input-level representations influence such gaps has not been investigated, particularly with phonemic inputs. 
We hypothesize that the performance gaps are affected by representation discrepancies between these languages, and revisit the use of phonemic representations as a means to mitigate these discrepancies.
To demonstrate the effectiveness of phonemic representations, we present experiments on three representative cross-lingual tasks on 12 languages in total. The results show that phonemic representations exhibit higher similarities between languages compared to orthographic representations, and it consistently outperforms grapheme-based baseline model on languages that are relatively low-resourced.
We present quantitative evidence from three cross-lingual tasks that demonstrate the effectiveness of phonemic representations, and it is further justified by a theoretical analysis of the cross-lingual performance gap.

\end{abstract}

%% file: sec/1_intro.tex
\section{Introduction} \label{sec:intro}

Large language models have significantly advanced natural language processing, offering improved capabilities across numerous languages. However, substantial \textbf{performance gaps} remain, particularly between high-resource languages like English and the majority of the world's low-resource languages. While these gaps are partly driven by discrepancies in data availability and quality, recent studies suggest that \textbf{linguistic gaps}—potentially caused by structural and lexical differences—also contribute significantly to these disparities .

Cross-lingual transfer techniques, which aim to adapt to arbitrary target language, have shown promise with the advancement of pre-trained multilingual language models \cite{devlin2018bert, conneau2020unsupervised, clark-etal-2022-canine}. However, they continue to face challenges, particularly with low-resource languages. One line of prior research has focused on mitigating these gaps through cross-lingual representation alignment \cite{zhang2022mixed, wu-monz-2023-beyond, Stap2023ViewingKT}, but these efforts often overlook the impact of varying input representations on performance consistency across languages.

\begin{figure}[t]
    \centering
    \includegraphics[width=0.49\textwidth]{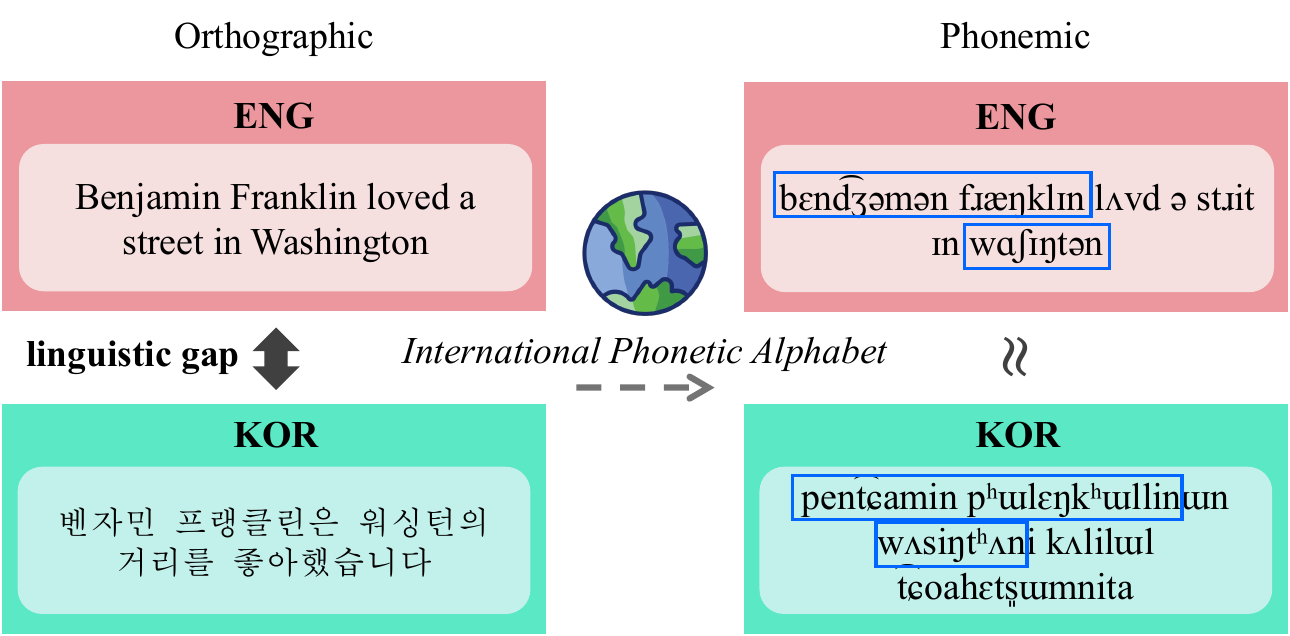}
    \caption{
    Example of orthographic and phonemic input representations of a sentence (English and Korean).}
    \label{fig:en_hi_ex}
\end{figure}

In this work, we explore the use of phonemic representations written in International Phonetic Alphabet (IPA) characters as a robust input representation (see Figure \ref{fig:en_hi_ex}) to reduce linguistic gaps and, consequently, performance gaps across languages.
We define the \textit{linguistic gap} as the representation discrepancy between embedding vectors and the \textit{performance gap} as the relative difference in downstream task performances between languages, to analyze the impact of phonemic representations in cross-lingual adaptation.

Our empirical analysis shows that phonemic representations consistently reduce linguistic gaps between languages compared to orthographic character-based models. This reduction in linguistic gaps directly correlates with smaller performance gaps in tasks such as cross-lingual natural language inference (XNLI), named-entity recognition (NER), and part-of-speech (POS) tagging, demonstrating the potential of phoneme-based models to enhance cross-lingual transfer across diverse languages. We further support these findings with theoretical analysis from domain generalization literature, where we frame the performance gap as a consequence of linguistic gaps driven by lexical and syntactic differences.

Our key contributions are as follows:
\begin{itemize}
    \item We revisit the use of phonemic representations (IPA) as a universal input strategy to reduce performance gaps across languages in multilingual language models.
    \item We empirically demonstrate the effectiveness of phonemic representations by comparing them with subword and character-based models, highlighting their ability to minimize both performance and linguistic gaps.
    \item We provide a theoretical explanation for the observed benefits of phonemic representations, drawing parallels between linguistic gaps in multilingual settings and domain gaps in domain generalization literature.
\end{itemize}

%% file: sec/2_related_works.tex
\section{Related Works} \label{sec:related_works}
\subsection{Cross-lingual Transfer with Multilingual Language Model}
Cross-lingual transfer learning aims to improve performance on low-resource languages (LRLs) by leveraging data from high-resource languages. Models like mBERT \cite{devlin2018bert} and XLM-R \cite{conneau2020unsupervised}, trained on hundreds of languages, have demonstrated effective cross-lingual adaptation by leveraging large multilingual pre-train datasets \cite{fujinuma-etal-2022-match, wu-dredze-2020-languages, conneau2020unsupervised}. However, significant performance discrepancies remain between languages due to differences in data availability, script types, and language families \cite{wu-dredze-2020-languages, muller-etal-2021-unseen, bagheri-nezhad-agrawal-2024-drives}. This "performance gap" has been systematically evaluated in benchmarks such as XTREME \cite{hu2020xtreme}, highlighting the need for methods that can ensure more consistent performance across languages.

\subsection{Cross-lingual Representation Gap}
One approach to reducing performance gaps focuses on narrowing the representation gap between languages. Multilingual pre-training enables models to learn shared representation space for multiple languages. 
\cite{singh-etal-2019-bert} and \cite{muller-etal-2021-first} both analyze the representations of pre-trained multilingual models and observe that lower layers are responsible for this cross-lingual alignment.
\cite{yang2022enhancing} employs mixup \cite{zhang2018mixup} to bring representations closer together, improving performance by reducing their distance in the latent space. Other works show a strong correlation between representation distance and machine translation performance, suggesting that improved alignment leads to better transfer results \cite{wu-monz-2023-beyond, Stap2023ViewingKT}.
While these studies provide valuable insights into the benefits of aligning cross-lingual representations, they do not explore how variations in input-level representations, such as the use of phonemic representations instead of orthographic characters, might affect this alignment. This paper investigates how phonemic representations can further reduce cross-lingual gaps.

\subsection{Phonemic Representations for Multilingual Language Modeling}
Phonemes, typically represented by International Phonetic Alphabet (IPA) characters, are the perceptual sounds of a language.
Phonemic representations offer a language-agnostic input that can enhance multilingual modeling, especially for LRLs. By using phonological features that are less dependent on specific orthographic systems, these representations offer a language-agnostic alternative that can help bridge performance gaps across languages. 
Previous studies have shown that using the IPA characters as input can enhance performance in cross-lingual tasks such as named entity recognition \cite{chaudhary2018adapting, bharadwaj2016phonologically, leong2022phone} and machine translation \cite{chaudhary2018adapting, sun-etal-2022-alternative}, particularly for low-resource languages.
Similarly, \citet{sohn2024zeroshotcrosslingualnerusing} report that phoneme-based models outperform other baselines on target languages unseen during pre-training. 
While these works demonstrate the potential of phonemic representations in language modeling, few have explored the specific embeddings and representations of phonemes.
Although some studies have developed pre-defined phoneme embeddings (e.g., PanPhon \cite{mortensen-etal-2016-panphon}, Phoible \cite{phoible}) and learned embeddings from masked language modeling \cite{plbert, pngbert, sundararaman2021phoneme, zhang2022mixed}, there is limited understanding of how these embeddings function in cross-lingual contexts.

We utilize XPhoneBERT \cite{nguyen2023xphonebert}, a model pre-trained with phonemes across approximately 100 languages, to investigate how using phonemic representations as input can mitigate cross-lingual performance discrepancies. Our empirical and theoretical analyses provide new insights into the benefits of phonemic representations for multilingual language modeling, particularly in terms of narrowing the cross-lingual linguistic gap and performance gap.

%% file: sec/3_experiments.tex
\section{Experimental Setup}
\label{sec:setup}
In this section, we describe the experiment setup in terms of models, datasets, and downstream tasks, including the selected target languages and details for preprocessing. Additionally, we provide details on evaluation strategies, particularly on quantifying the performance and linguistic gap.

\subsection{Models}
\label{subsec:models}
We employ three masked language models that are pre-trained on multilingual corpus that covers around 100 languages from Wikipedia dump files\footnote{pre-trained weights are obtained from https://huggingface.co/models}: mBERT \cite{devlin2018bert}, CANINE \cite{clark-etal-2022-canine}, and XPhoneBERT \cite{nguyen2023xphonebert}. Each model is trained on different types of language representation.

Multilingual BERT (mBERT) is a \textbf{subword-based} model that utilizes WordPiece algorithm for tokenization. During pre-training, mBERT learns to perform masked language modeling (MLM) and next sentence prediction (NSP).

CANINE is a multilingual \textbf{character-based} model that is trained on the same corpus with the same training objective as mBERT. CANINE is a tokenization-free language model that directly maps each unicode character to its codepoint by hashing. This prevents unknown tokens, enabling the model to handle a large amount of distinct characters.

Lastly, XPhoneBERT is a \textbf{phoneme-based} model trained to do MLM. XPhoneBERT follows the pre-training scheme of XLM-R \cite{conneau2020unsupervised}, so NSP is not employed in its pre-training. This model takes as input the sequence of IPA characters, where the input data are created from original text by G2P conversion followed by phoneme segmentation.

While character-level models are known to better generalize to low-resource languages \cite{clark-etal-2022-canine}, their general performance falls behind subword-based models. 
To specifically compare input representations--phonemes versus orthographic scripts--we minimize the impact of different tokenization units by focusing on phoneme-based models versus character-based models, rather than directly comparing with subword-based models like mBERT. Nevertheless, we include mBERT results for the XNLI task to highlight its significant performance drop on low-resource languages. For other tasks, we report results from phoneme and character-based models to ensure a fair comparison, and leave further improvements of character-level models in overall performance as future work.

\subsection{Downstream Tasks}
We adopted the cross-lingual generalization benchmark tasks suggested in XTREME \cite{hu2020xtreme}.

\paragraph{Token-level Classification.} We choose \textbf{POS tagging} and \textbf{NER} as our testbed for structured prediction tasks. Both tasks require labeling each token from the model. These types of tasks were previously analyzed as being relatively independent from the data size of each language used for pre-training the language model \cite{hu2020xtreme}. We find this particularly suitable in our scenario where two models with different pre-training strategy are compared. For datasets, we utilize the corpora from Universal Dependencies\footnote{https://universaldependencies.org/ , v2.13, 148 languages, released Nov 15, 2023.} for POS tagging, and WikiAnn \cite{pan-etal-2017-cross} with train, dev, test splits following \citet{rahimi2019massively} for NER.

\paragraph{Sentence-level Classification.} 
XTREME supports two sentence-level classification tasks. This  type of task requires semantic understanding of given sentences to make a prediction. We employ \textbf{XNLI} \cite{conneau2018xnli} dataset, which is a representative benchmark for the natural language inference task on cross-lingual generalization setting.
This task requires the model to classify the relation of two given sentences into three different classes. 

\subsection{Performance Gap}
\label{sec:rpd}
We analyze performance gaps of each model for all downstream tasks. As we are interested in how different models with different input types performs consistent across languages rather than their absolute overall performance, we take the relative percentage difference (RPD) \cite{rpd} to derive the performance gap. Here, we define RPD as
\begin{equation}
    \text{RPD}(L_i, L_j) = \frac{|\text{S}(L_{i}) - \text{S}(L_{j})|}{\frac{1}{2}(\text{S}(L_{i}) + \text{S}(L_{j}))} \times 100,
\end{equation}
where S$(L_i)$ represents the performance for the language $L_i$. This is used to analyze the performance gap, which specifically computes the relative performance gaps across languages.

\subsection{Linguistic Gap}
To compute representation discrepancy across languages, we use FLORES+ \cite{nllb-22} corpus which contains parallel sentences of more than 200 languages. We employ devtest set of each language subset, which contains 1,012 sentences. 

After training each model on each downstream task, we utilize each model to obtain similarity in their representations. We adopt mean-pooling to obtain sentence representations and Centered Kernel Alignment (CKA) \cite{cka} to measure the similarity, which \citet{Del2021SimilarityOS} has recommended for robust analysis on cross-lingual similarity. CKA is defined as,
\begin{equation}
    \text{CKA}(\mathbf{X}, \mathbf{Y}) = \frac{\| \mathbf{X}^T \mathbf{Y} \|_2^2}{\| \mathbf{X}^T \mathbf{X} \|_2 \| \mathbf{Y}^T \mathbf{Y} \|_2},
\end{equation}
where features $\mathbf{X}$ and $\mathbf{Y}$ are from different languages. They are extracted from the input embedding layers as we are interested in how different input types (i.e., orthographic vs. phonemic) affect cross-lingual alignment, and \citet{muller-etal-2021-first} finds that cross-lingual alignment happens in the lower layers of the model. 
We use this similarity scores computed with CKA to refer to \textit{linguistic gaps}, where smaller CKA score means larger linguistic gap.

\subsection{Implementation Details}
Models were trained for 30 epochs on a single NVIDIA A5000 GPU for POS tagging, 30 epochs a single NVIDIA A40 GPU for NER, and 20 epochs on NVIDIA A6000 for XNLI. 
For all experiments, batch size was set to 128 and AdamW \cite{loshchilov2018decoupled} optimization was used. Additionally, cosine learning rate scheduler was adopted with its initial learning rate set by grid search. Learning rates used for each model on each language are in the supplementary material.

\subsection{Data Preparation}
\paragraph{Languages.}
To evaluate token-level tasks, we selected 10 languages with diverse typoloigical background---English(\texttt{eng}), French(\texttt{fra}), Russian(\texttt{rus}), Italian(\texttt{ita}), Hungarian(\texttt{hun}), Ukrainian(\texttt{ukr}), Korean(\texttt{kor}), Turkish(\texttt{tur}), Finnish(\texttt{fin}), and Hindi(\texttt{hin}).
First four languages are high-resource languages, where English, French, and Italian are written in Latin scripts and Russian in Cyrillic. 
The other languages are pre-trained on each model with moderate or small amount of data, and are written in diverse scripts, such as Hangul, Cyrillic and Devanagari. For further analysis using sentence-level tasks, we chose two low-resource languages---Swahili(\texttt{swa}), and Urdu(\texttt{urd})---to compare with a representative high-resource language, English(\texttt{eng}).

\paragraph{Preprocessing.} In order to prepare inputs for a phoneme-based model, we employed G2P (Grapheme-to-Phoneme) conversion to obtain an IPA version of the input. This conversion was done with Epitran\footnote{https://github.com/dmort27/epitran} \cite{mortensen-etal-2018-epitran}
, an external tool for G2P conversion. After converting to IPA, phoneme segmentation with a python package, segments\footnote{https://pypi.org/project/segments/}, to identify each phoneme. Lastly, to make it compatible with XPhoneBERT's tokenizer, white space was inserted between every phoneme.

%% file: sec/4_numerical_analysis.tex
\section{Results and Analysis} \label{sec:numerical_analysis}

Here, we present our observations and analyses of the results. We first discuss the behavior of phoneme-based model towards low-resource languages and writing systems, which contributes to robust cross-lingual performance. Next, we delve into the performance and linguistic gaps of phoneme-based models through empirical and theoretical analyses.

\input{tables/main_table}

\subsection{Phoneme-based Model on Low-Resource Languages and Writing Systems} 
We observe that phoneme-based model shows promising performance in low-resource languages and writing systems (scripts). Results from Table \ref{tab:main_structpred} show that phoneme-based model outperforms the character-based model on NER task, in languages written in scripts other than major scripts\footnote{Latin and Cyrillic are scripts that are used the most during the pre-training phase.}---Korean and Hindi. This can be attributed to the fact that named entities, such as geopolitical or personal names, are often pronounced similarly across languages. When different writing systems and scripts are used, models may struggle to align such entities. However, representing them in IPA characters that reflect their pronunciations helps the model to better align these entities, resulting in better cross-lingual transfer. This results align with findings from \citet{muller-etal-2021-unseen, sohn2024zeroshotcrosslingualnerusing}, which focus on unseen languages, whereas we observe this phenomenon with diverse `seen' languages.

Results also demonstrates the potential of phoneme-based model in addressing low-resource languages. As shown in Table \ref{tab:xnli}, the phoneme-based model achieves a smaller gap when transferred to low-resource languages such as Swahili and Urdu, compared to other baselines. This finding is further analyzed in Section \ref{subsec:perf-gap}

\subsection{Performance Gap Across Languages}
\label{subsec:perf-gap}

\begin{table}[th]
\centering\small
\resizebox{\columnwidth}{!}{
\begin{tabular}{@{}l|c|cc|cc@{}} 
\toprule
\multirow{4}{*}{Method}    & \multicolumn{5}{c}{Language} \\ \cmidrule{2-6}
          & \texttt{eng} & \multicolumn{2}{c}{\texttt{swa}}  & \multicolumn{2}{|c}{\texttt{urd}} \\ \cmidrule{2-6}
          & \multirow{2}{*}{Acc.} & \multirow{2}{*}{Acc.} & $\Delta$ from \texttt{eng} & \multirow{2}{*}{Acc.} & $\Delta$ from \texttt{eng} \\
          &&& (Rel./Abs.) && (Rel./Abs.) \\
          \midrule
Subword & 80.80 & 62.93 & 24.87 / 17.87 & 61.57 & 27.01 / 19.23 \\
Character & 75.02 & 59.72 & 22.71 / 15.30 & 56.55 & 28.08 / 18.47 \\
Phoneme & 71.89 & 60.88 & \textbf{16.59} / \textbf{11.01} & 56.10 & \textbf{24.67} / \textbf{15.79} \\
\bottomrule
\end{tabular}}
\caption{Accuracy (\%) and relative/absolute performance gaps on XNLI task. \texttt{eng}, \texttt{swa}, and \texttt{urd} refer to English, Swahili, and Urdu, respectively, and relative difference is computed with RPD. Phonemic representation shows relatively small performance gaps compared to other representations.}
\label{tab:xnli}
\end{table}

\begin{figure*}[!t]
    \centering
    \includegraphics[width=0.9\linewidth]{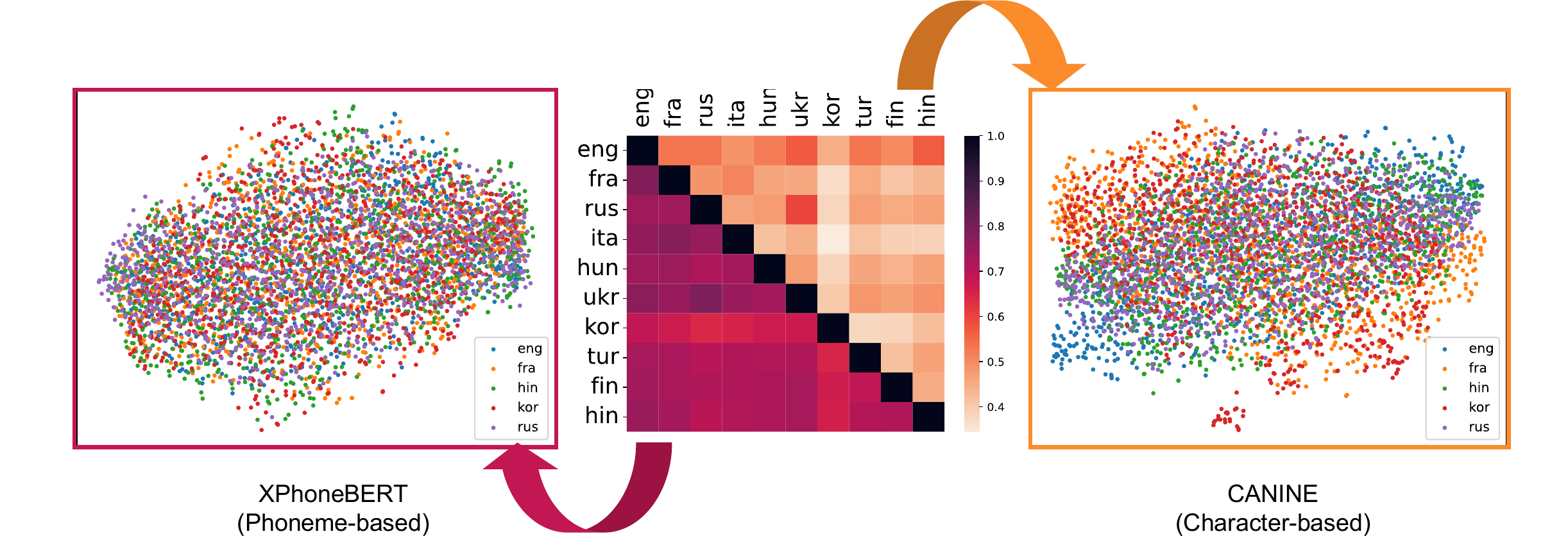}
    \caption{Linguistic gaps across languages in each model. (Center) Upper and lower triangular elements of the heatmap indicate pairwise linguistic gaps derived with character-based model and phoneme-based model, respectively. Darker color indicates larger CKA score, which means smaller discrepancy. Lower triangular elements show relatively darker colors, implying smaller discrepancies across languages of phoneme-based model. (Left, right) T-SNE plots for each model are shown with only five languages, for better visibility.}
    \label{fig:heatmap}
\end{figure*}

We observe that the phoneme-based model consistently exhibits the smallest performance gap across diverse languages, highlighting its robustness in cross-lingual tasks.
In Table \ref{tab:main_structpred}, we present the standard deviation (Std.) and average percentage difference (Mean diff.) for all models, which reflect the variability in performance across different languages.
The phoneme-based model exhibits both a lower standard deviation and a smaller average percentage difference in the NER and POS tasks, demonstrating its relatively stable performance across different languages. 

Table \ref{tab:xnli} provides additional evidence by showing that the phoneme-based model achieves a smaller gap in performance between English and other low-resource languages---Swahili(\texttt{swa}) and Urdu(\texttt{urd})---compared to other models. We report both relative and absolute differences in performance, with the relative difference calculated as described in Section \ref{sec:rpd}.

\begin{figure*}[!h]
    \centering
    \includegraphics[width=0.9\textwidth]{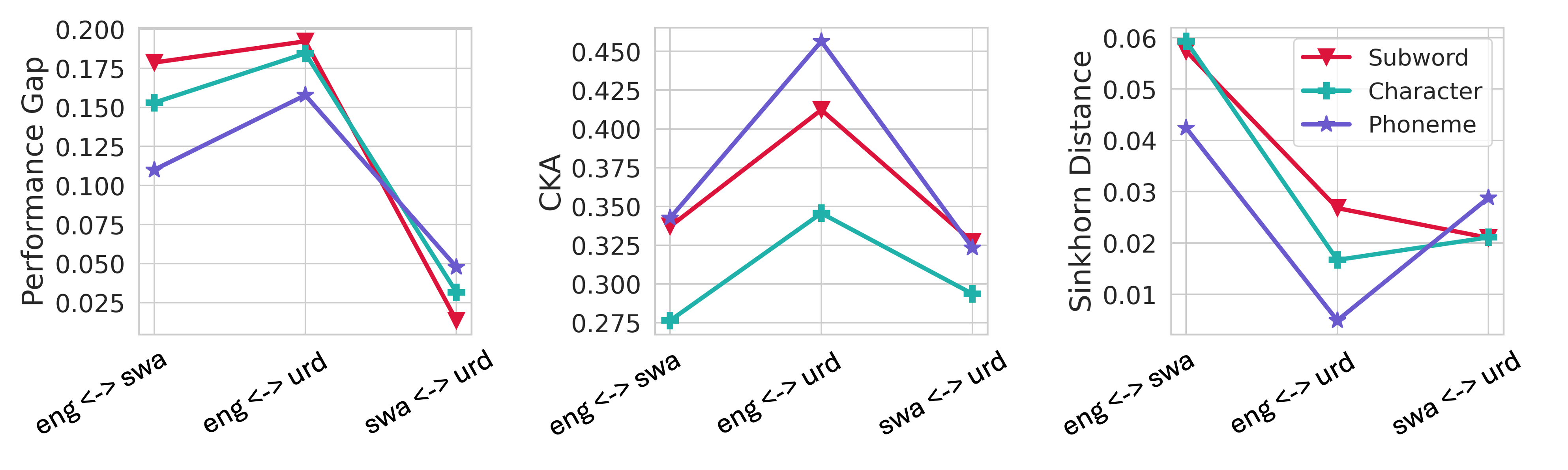}
    \caption{Qualitative analysis of performance gap (difference of accruacy) on XNLI task. (Left) the absolute difference between performance across two languages, (center) centered kernel alignment (CKA) scores to measure cross-lingual embedding similarity, and (right) Sinkhorn distance on the output probability space. Phonemic representation shows relatively small performance gaps w.r.t. \texttt{eng} $\leftrightarrow$ \texttt{swa} and \texttt{eng} $\leftrightarrow$ \texttt{urd}, and these gaps are correlated with similarity and discrepancy on the embedding space (CKA) and logit space (Sinkhorn distance).}
    \label{fig:qualitative analysis}
\end{figure*}

While subword-based mBERT achieves the highest scores, the performance gaps between models narrow when applied to low-resource languages, with outperforming the phoneme-based model by 8.91\% in English and by 2.05\% and 5.47\% in Swahili and Urdu, respectively. This reflects subword LM's significant performance drops on low-resource languages, while highlighting the phoneme-based LM's robustness in cross-lingual transfer to such languages.
The leftmost panel of Figure \ref{fig:qualitative analysis} also illustrates the performance gaps of each model, where the phoneme-based model predominantly displays lower gaps compared to others. 

These metrics collectively suggest that phonemic representations offer a more consistent performance in multilingual settings, reducing the disparities typically observed when models are applied to languages with varying resource availability.

\subsection{Linguistic Gap of Different Representations} 
\label{subsec:ling-gap}
To investigate the potential of phonemes as a robust representation for multilingual language modeling, we analyze the linguistic gap between languages using different input representations. Following \citet{yang2022enhancing, muller-etal-2021-first}, we use linear CKA to quantify representation similarity across languages. Figure \ref{fig:heatmap} shows the pairwise similarities between languages, with the lower triangle of the heatmap, which corresponds to phonemic representations, demonstrating higher similarity values. This indicates a smaller linguistic gap compared to models that use orthographic inputs, contributing to a smaller performance gap. Moreover, the t-SNE plots placed in both sides show how the distributions of the representations from different languages resemble each other. Phoneme-based model exhibits more similar distribution across languages.

Figure \ref{fig:qualitative analysis} further supports these observations by showing the linguistic gap after fine-tuning on the XNLI task. The plot in the center illustrates that phonemic representations have higher CKA scores than other baseline models, indicating closer alignment between language representations. 
As XNLI directly learns to build a sentence representation during fine-tuning, we extract the representation from the last hidden layer unlike in other token-level tasks.
Additionally, by using Sinkhorn distance to compare the logit space, we observe that the phoneme-based model shows lower distances, reflecting more consistent predictions across languages.
\bigskip

These results highlight the potential of phonemic representations to address the performance gaps that challenge multilingual language models, particularly in bridging the gap between high-resource and low-resource languages by more similar representations. 

\subsection{Connecting Performance Gap and Linguistic Gap}

\begin{table*}[!ht]
    \centering
    \resizebox{0.9\linewidth}{!}{
    \begin{tabular}{l|cc|cc}
    \toprule
        \multirow{2}{*}{Correlation} & \multicolumn{2}{c|}{Spearman's R} & \multicolumn{2}{c}{Kendall's T} \\\cmidrule{2-5}
         & coefficient & p-value & coefficient & p-value \\ \midrule
        Performance Gap  <-> \texttt{eng} Performance & 0.111 & 5.60E-01 & 0.104 & 4.30E-01 \\ 
        \midrule
        Performance Gap <-> \textbf{S-Dist} & \textbf{0.681} & \textbf{3.50E-05} & \textbf{0.457} & \textbf{2.00E-04} \\ 
         \midrule
        Performance Gap <-> \textbf{CKA}   & \textbf{-0.782} & \textbf{3.40E-07} & \textbf{-0.577} & \textbf{2.10E-06} \\\bottomrule
    \end{tabular}}
    \caption{\label{tab:correlation_anal}Correlation analysis with 45 phoneme-based models. We fine-tune the phoneme-based language model XPhoneBERT on three languages, \texttt{eng}, \texttt{swa}, and \texttt{urd}, with 15 different random seeds and conduct two types of correlation analyses.}
\end{table*} 

\paragraph{Correlation Analysis.} Meanwhile, one may speculate the low-performance gap of the phoneme-based model can be driven by the low English performance rather than reducing the linguistic gap. To clarify this, we simulate 15 repeated runs (with different random seeds) of phonemic representation using 10\% of the XNLI train dataset over English, Swahili, and Urdu. After computing the best performance per each language, Sinkhorn distance (S-Dist), and CKA between English and the other two languages, we conducted correlation analyses by performing hypothesis tests with Spearman's rank correlation coefficient and Kendall's Tau. 

As can be seen from Table \ref{tab:correlation_anal}, rather than the English performance, S-Dist and CKA have stronger correlations, indicating that the linguistic gap has stronger correlations that are statistically significant (with a significant level less than 0.01).

%% file: tables/main_table.tex
\setlength{\tabcolsep}{4pt}
\renewcommand{\arraystretch}{0.9} 
\begin{table*}[t]
	\begin{center}
    	\resizebox{\linewidth}{!}{
            \begin{tabular}{@{}l|cccccccccc|c|c|c@{}}
            \toprule
            \multirow{2}{*}{Method} & \multicolumn{10}{c|}{Language} & \multicolumn{2}{c|}{Performance gap} & Linguistic Gap \\\cmidrule{2-14}
                                    & \texttt{eng} & \texttt{fra} & \texttt{rus} & \texttt{ita} & \texttt{hun} & \texttt{ukr} & \texttt{kor} & \texttt{tur} & \texttt{fin} & \texttt{hin} & Std. ($\downarrow$) & Mean RPD ($\downarrow$) & Mean CKA ($\uparrow$)\\\midrule
            \multicolumn{14}{c}{\textit{Named Entity Recognition}}\\\midrule
            Character &  87.13 & 91.27 & 91.80 & 92.26 & 93.14 & 93.88 & 84.11 & 92.92 & 90.45 & 87.68 & 0.0316 & 4.02 & 0.4584 \\ 
            Phoneme &  83.61 & 89.42 & 89.60 & 90.56 & 91.89 & 92.76 & 87.19 & 92.35 & 89.23 & 88.23 &  \textbf{0.0259} & \textbf{3.52} & \textbf{0.7195} \\\midrule
            \multicolumn{14}{c}{\textit{Part-of-Speech Tagging}}\\ \midrule
            Character & 96.62 & 95.54 & 87.91 & 96.06 & 74.57 & 85.79 & 86.71 & 90.49 & 91.78 & 96.81 & 0.0692 & 8.77 &  0.4593\\ 
            Phoneme & 95.94 & 96.35 & 86.69 & 96.37 & 85.87 & 91.32 & 85.82 & 91.11 & 93.76 & 96.94 & \textbf{0.0455} & \textbf{5.80} & \textbf{0.7204} \\\bottomrule
            \end{tabular}}
     \end{center}
     \caption{Performance of POS tagging and NER across different languages. Std. refers to the standard deviation of the scores across the languages, and Mean RPD indicates average relative difference of F1 scores between different languages. Mean CKA represents the average linguistic gap between languages.}
     \label{tab:main_structpred}
\end{table*}

%% file: sec/5_theoretical_analysis.tex
\paragraph{Theoretical Analysis.}

We aim to diminish the performance gap between different languages by adopting IPA as a universal language representation. Motivated by domain adaptation literature \cite{kifer2004, bendavid2010}, we present a theoretical justification of IPA for robust multilingual modeling by deriving a bound for cross-lingual performance gap. 

Let $\mathcal{D}$ denote a domain as a distribution over text feature input $\mathcal{X}$, such as the sequence of word embeddings or one-hot vectors, and a labeling function $f:\mathcal{X} \rightarrow [0,1]$. Assuming a binary classification task, our goal is to learn a hypothesis $h:\mathcal{X} \rightarrow \{0,1\}$ that is expected to minimize a risk $\varepsilon_{D}(h, f):=\mathbb{E}_{x \sim \mathcal{D}}[|f(x) - h(x)|]$ and has a small risk-deviation over two domains $\mathcal{D}_{A}$ and $\mathcal{D}_{B}$. Then, to formalize the cross-lingual performance gap, we first need a discrepancy measure between two languages. By following \citet{bendavid2010}, we adopt $\mathcal{H}$-divergence (See Appendix \ref{sec:apx:thm} for its definition) to quantify the distance between two language distributions. 

Now, based on Lemma 1 and 3 of \citet{bendavid2010}, we make reasoning on performance gap over different language domains.
\begin{theorem} \label{thm:bound}
Let $h:\mathcal{X} \rightarrow \{0,1\}$ be a function in a hypothesis class $\mathcal{H}$ with a pseudo dimension $\mathcal{P} dim(\mathcal{H})=d$. If $\hat{\mathcal{D}}_{A}$ and $\hat{\mathcal{D}}_{B}$ are the empirical distribution constructed by $n$-size i.i.d. samples, drawn from $\mathcal{D}_{A}$ and $\mathcal{D}_{B}$ respectively, then for any $\delta \in (0,1)$, and for all $h$, the bound below hold with probability at least $1-\delta$.

\begin{align}
|\varepsilon_{\mathcal{D}_{A}}(h, f) - \varepsilon_{\mathcal{D}_{B}}(h, f)| \le \frac{1}{2} d_{\mathcal{H}\Delta\mathcal{H}}(\hat{\mathcal{D}}_{A}, \hat{\mathcal{D}}_{B}) \nonumber \\
\nonumber
     + 2 \sqrt{\frac{d \log(2n) + \log(2/\delta)}{n}}
\end{align}
\end{theorem}

\noindent 
where $\mathcal{H}\Delta\mathcal{H}:=\{h(x) \oplus h'(x)|h,h' \in \mathcal{H}\}$ given $\oplus$ as a \texttt{xor} operation (proof is in Appendix \ref{sec:apx:thm}). We see that performance gap between two languages is bounded from above with a distribution divergence plus an irreducible term defined by problem setup. That is, if we reduce the divergence between language distributions, the expected performance gap can also be reduced accordingly. 

To investigate whether this is indeed a case or not, we provided embedding space similarity and logit-space Sinkhorn distance~\cite{sinkhorn} between different languages in Figure \ref{fig:qualitative analysis}. We argue that phonemic representation's relatively mild performance gap is achieved by reducing linguistic gaps which is confirmed in the embedding space (high CKA) and final output space (low Sinkhorn distance). 

%% file: sec/6_conclusion_limitation.tex
\section{Conclusion} \label{sec:conclusion}
Towards robust multilingual language modeling, we argue that mitigating the linguistic gap between different languages is crucial. Moreover, we advocate the use of IPA phonetic symbols as a universal language representation partially bridges such linguistic gaps without any complicated cross-lingual training phase. Empirical validation on three representative NLP tasks demonstrates the superiority of phonemic representation compared to subword and character-based language representation in terms of the cross-lingual performance gap and linguistic gap. Theoretical analysis of the cross-lingual performance gap explains such promising results of phonemic representation.

\section{Limitations} While we have shown that phonemic representation induces a small cross-lingual linguistic gap, therefore a small performance gap, the absolute performance of this phonemic representation is still lacking compared to subword-level models. We spur the necessity of putting research attention to developing phoneme-based LMs. Moreover, there is no such large phonemic language model beyond the BERT-base-size architecture, so we confine the scope of our empirical validation to BERT-base-size LMs. 
This also means the experiments rely on existing pre-trained models, limiting control over their pre-training settings. Since the models were trained on different language sets and pre-training objectives (as noted in \ref{subsec:models}), it is important to verify these findings in a controlled environment.
Additionally, we performed evaluation with a limited languages (up to 12), so it is unclear whether IPA language representations are effective for other numerous languages (especially low-resource ones) or not.

%% file: sec/10_impact.tex
\section{Impact Statement}
Ensuring strong multilingual understanding capability in our language modeling systems is increasingly crucial given the openness of frontier-level AI produce \cite{dubey2024llama}. Enhanced cross-lingual transferability by leveraging our findings may have some positive impacts on diverse domains such as healthcare \cite{he2023survey,ryu-etal-2024-ehr} and education \cite{han2023recipe,OH2024123960} which are very close to our daily life as well as on further development of cutting-edge multimodal large language models \cite{kim2023visually, yin2023survey}.

%% file: sec/7_ethics.tex
\section{Ethics Statement}
We believe there are no potential of any critical issues that harm the code of ethics provided by ACL. The social impacts of the technology---reducing performance gaps for low resource languages---will be, on the balance, positive. The data was, to the extent we can determine, collected in accordance with legal and institutional protocals.

%% file: sec/8_acknowledgement.tex
\section*{Acknowledgements}
\label{sec:acknowledge}
We sincerely thank Haneul Yoo at KAIST who provided constructive feedback on the manuscript. This work was supported by Institute of Information \& communications Technology Planning \& Evaluation (IITP) grant funded by the Korea government(MSIT) (RS-2022-00143911, AI Excellence Global Innovative Leader Education Program)

%% file: sec/9_appendix.tex
\newpage

\section{Dataset Statistics}
In Table \ref{tab:dataset_stats}, we provide the dataset statistics. For the experiments, we used train set for training and validation set for evaluation.

\begin{table}[h]
    \centering
    \scriptsize
    \begin{tabular}{c|c|c|c|c}
    \toprule
         Dataset & Lang. & Train & Dev & Test \\ \midrule
         \multirow{10}{*}{FLORES+} & \texttt{eng} & \multirow{10}{*}{-} & \multirow{10}{*}{1.2k} & \multirow{10}{*}{-}  \\
                                &\texttt{fra} & & &   \\ 
                                &\texttt{rus} & & &   \\ 
                                &\texttt{ita} & & &   \\ 
                                &\texttt{hun} & & &   \\ 
                                &\texttt{ukr} & & &   \\ 
                                &\texttt{kor} & & &   \\ 
                                &\texttt{tur} & & &   \\ 
                                &\texttt{fin} & & &   \\ 
                                &\texttt{hin} & & &   \\\midrule
          \multirow{3}{*}{XNLI} & \texttt{eng} & \multirow{3}{*}{393k} & \multirow{3}{*}{2.49k} & \multirow{3}{*}{5.01k}  \\ 
                                &\texttt{swa} & & &   \\ 
                                & \texttt{urd} & & &   \\\midrule
         \multirow{8}{*}{WikiAnn} & \texttt{eng}& 20k & 10k & 10k  \\ 
                                & \texttt{fra}& 20k & 10k & 10k   \\
                                & \texttt{rus}& 20k & 10k & 10k   \\
                                & \texttt{ita}& 20k & 10k & 10k   \\
                                & \texttt{hun}& 20k & 10k & 10k   \\
                                & \texttt{ukr}& 20k & 10k & 10k  \\
                                & \texttt{kor}& 20k & 10k & 10k  \\
                                & \texttt{tur}& 20k & 10k & 10k  \\
                                & \texttt{fin}& 20k & 10k & 10k  \\
                                & \texttt{hin}& 5k & 1k & 1k \\
                                \midrule
        \multirow{6}{*}{UD} & \texttt{eng}& 12.5k & 2k & 2k \\ 
                            & \texttt{fra}& 14.5k & 1.5k & 0.4k \\ 
                            & \texttt{rus} & 16k & 0.9k & 0.9k \\
                            & \texttt{ita} & 13k & 0.6k & 0.5k \\
                            & \texttt{hun} & 0.9k & 0.4k & 0.4k \\
                            & \texttt{ukr} & 5.5k & 0.7k & 0.9k \\
                            & \texttt{kor}& 23k & 2k & 2.3k \\ 
                            & \texttt{tur} & 15k & 1.6k & 1.6k \\
                            & \texttt{fin}& 12k & 1.4k & 1.6k \\
                            & \texttt{hin} & 13k & 1.7k & 1.7k \\ 
         \bottomrule

  \end{tabular}
  \caption{Dataset statistics for datasets used in experiments: FLORES+, XNLI, WikiAnn, Universal Dependencies Tree Bank. For FLORES+ dataset, we used devtest set with 1,012 sentences.}

    \label{tab:dataset_stats}
\end{table}

\section{Hyperparameter sweep.}
\begin{table*}[]
    \centering
    \scriptsize
    \resizebox{\linewidth}{!}{
    \begin{tabular}{c|c|c|c|c|c}
        \toprule
        \multirow{2}{*}{Task} & \multirow{2}{*}{Hyperparam} & \multirow{2}{*}{Search space} & \multicolumn{3}{c}{Selected parameter value} \\ \cmidrule{4-6}
                                &           &               & mBERT &CANINE&XPhoneBERT \\ \midrule
        \multirow{3}{*}{XNLI} & learning rate & [5e-6, 7e-6, 1e-5, 3e-5, 5e-5] & 5e-6 & 5e-6 (en), 1e-5 (sw, ur) & 7e-6 (en), 3e-6 (sw, ur) \\ 
                & weight decay &  [0.0, 1e-1, 1e-2, 1e-3] & 0.01 & 0.1 (en), 0.0 (sw), 0.01 (ur) & 0.1 (en), 0.0 (sw), 0.01 (ur) \\ 
                & learning rate scheduling &  [True, False] & True & True & False \\ \midrule
        \multirow{2}{*}{NER} & learning rate & [3e-5, 5e-5, 1e-4, 3e-4] & - & 5e-5 (en, fr, it, hu, ko, tr), 1e-4 (ru, uk, fi, hi) & 3e-5 (ru, it), 5e-5 (en, fr, hu, uk, tr, fi, hi), 1e-4 (ko) \\ 
              & weight decay & 1e-2 & - & 1e-2 & 1e-2 \\ \midrule
        \multirow{2}{*}{POS} & learning rate & [3e-5, 5e-5, 1e-4, 3e-4] & - & 5e-5 (ru, uk, tr), 1e-4 (en, fr, fi, hi), 3e-4 (it, hu, ko) & 5e-5 (en), 1e-4 (fr, ru, it, hu, uk, ko, tr, fi, hi) \\
          & weight decay & 1e-2  & - & 1e-2 & 1e-2 \\ \bottomrule

    \end{tabular}}
    \caption{List of hyperparameter, search spaces and selected parameter values for different models applied to XNLI, NER, and POS tasks, detailing learning rate, weight decay, and learning rate scheduling for mBERT, CANINE, and XPhonemBERT, with specific configurations for optimal model performance per task.}

    \label{tab:my_label}
\end{table*}

We sweep hyperparameters over grid below (in Table \ref{tab:my_label}), and select the final parameters for each model based on the \textbf{best validation performance} (Accuracy for XNLI and F1-score for NER and POS Tagging).


\newpage
\section{Details on Theoreoretical Analysis} \label{sec:apx:thm}

We aim to diminish the performance gap between different languages by adopting IPA as a universal language representation. Motivated by domain adaptation literature \cite{kifer2004, bendavid2010,oh2023towards}, we present a theoretical justification of IPA for robust multilingual modeling by providing a bound for cross-lingual performance gap.

Let $\mathcal{D}$ denote a domain as a distribution over text feature input $\mathcal{X}$, such as the sequence of word embeddings or one-hot vectors, and a labeling function $f:\mathcal{X} \rightarrow [0,1]$. Assuming a binary classification task, our goal is to learn a hypothesis $h:\mathcal{X} \rightarrow \{0,1\}$ that is expected to minimize a risk $\varepsilon_{D}(h, f):=\mathbb{E}_{x \sim \mathcal{D}}[|f(x) - h(x)|]$ and has a small risk-deviation over two domains $\mathcal{D}_{A}$ and $\mathcal{D}_{B}$. Then, to formalize the cross-lingual performance gap, we first need a discrepancy measure between two languages.  By following \cite{bendavid2010}, we adopt $\mathcal{H}$-divergence to quantify the distance between two language distributions.

\begin{definition}[$\mathcal{H}$-divergence; \citet{bendavid2006}]
Let $\mathcal{H}$ be a hypothesis class for input space $\mathcal{X}$ and a collection of subsets from $\mathcal{X}$ is denoted by $\mathcal{S}_\mathcal{H}:=\{h^{-1}(1)|h\in \mathcal{H}\}$ which is the support of hypothesis $h\in \mathcal{H}$. The $\mathcal{H}$-divergence between two distributions $\mathcal{D}$ and $\mathcal{D'}$ is defined as 
\begin{equation}
    d_{\mathcal{H}}(D,D')=2 \sup_{S\in \mathcal{S}_{\mathcal{H}}}|\mathbb{P}_{D}(S)-\mathbb{P}_{D'}(S)| \nonumber
\end{equation}
\end{definition}
\noindent 
$\mathcal{H}$-divergence is a relaxation of total variation between two distributions, and it can be estimated by finite samples from both distributions if $\mathcal{H}$ governs a finite VC dimension. Now, based on Lemma 1 and 3 of \citet{bendavid2010}, we make reasoning on performance gap over different language domains.

\begin{theorem} \label{thm:bound}
Let $h:\mathcal{X} \rightarrow \{0,1\}$ be a function in a hypothesis class $\mathcal{H}$ with a pseudo dimension $\mathcal{P} dim(\mathcal{H})=d$. If $\hat{\mathcal{D}}_{A}$ and $\hat{\mathcal{D}}_{B}$ are the empirical distribution constructed by $n$-size i.i.d. samples, drawn from $\mathcal{D}_{A}$ and $\mathcal{D}_{B}$ respectively, then for any $\delta \in (0,1)$, and for all $h$, the bound below hold with probability at least $1-\delta$.

\begin{align}
|\varepsilon_{\mathcal{D}_{A}}(h, f) - \varepsilon_{\mathcal{D}_{B}}(h, f)| \le \frac{1}{2} d_{\mathcal{H}\Delta\mathcal{H}}(\hat{\mathcal{D}}_{A}, \hat{\mathcal{D}}_{B}) \nonumber \\ \nonumber
     + 2 \sqrt{\frac{d \log(2n) + \log(2/\delta)}{n}}
\end{align}
\end{theorem}

\noindent 
where $\mathcal{H}\Delta\mathcal{H}:=\{h(x) \oplus h'(x)|h,h' \in \mathcal{H}\}$ given $\oplus$ as a \texttt{xor} operation.

\begin{proof}[proof of Theorem B.2.]
we start to prove Theorem B.2. by restating Lemma 1 of \cite{bendavid2010} adapted to our notation.

\begin{lemma} 
    Let $\mathcal{D}_{A}$ and $\mathcal{D}_{B}$ be distributions of domain $A$ and $B$ over $\mathcal{X}$, respectively. Let $\mathcal{H}$ be a hypothesis class of functions from $\mathcal{X}$ to $[0,1]$ with VC dimension d. If $\hat{\mathcal{D}}_{A}$ and $\hat{\mathcal{D}}_{B}$ are the n-size empirical distributions generated by $\mathcal{D}_{A}$ and $\mathcal{D}_{B}$ respectively, then, for $0< \delta <1$, with probability at least $1-\delta$, 
    \begin{align}
        d_{\mathcal{H}}(\mathcal{D}_{A}, \mathcal{D}_{B}) &\leq d_{\mathcal{H}}(\hat{\mathcal{D}}_{A}, \hat{\mathcal{D}}_{B}) \nonumber \\ \nonumber &+ 4 \sqrt{{{d \log(2n) + \log(2/\delta)} \over n}}.
    \end{align}
    \label{lem:div_bound}
\end{lemma}
\noindent
Then, for any hypothesis function $h, h' \in \mathcal{H}$, by the definition of $d_{\mathcal{H}\Delta\mathcal{H}}$-divergence, we have:

\small
\begin{align}
    & d_{\mathcal{H}\Delta\mathcal{H}}(\mathcal{D}_{A}, \mathcal{D}_{B}) \nonumber  \\
    &= 2 \sup_{h,h'\in \mathcal{H}}|\mathbb{P}_{x \sim \mathcal{D}_{A}}[h(x)\neq h'(x)]-\mathbb{P}_{x \sim \mathcal{D}_{B}}[h(x)\neq h'(x)]| \nonumber \\
    &= 2 \sup_{h,h'\in \mathcal{H}}|\varepsilon_{\mathcal{D}_{A}}(h,h') - \varepsilon_{\mathcal{D}_{B}}(h,h')| \nonumber \\
    & \geq 2 |\varepsilon_{\mathcal{D}_{A}}(h,h') - \varepsilon_{\mathcal{D}_{B}}(h,h')|  \nonumber
\end{align}
\normalsize
\noindent
Now the below bound holds for any hypothesis functions $h, h' \in \mathcal{H}$ (See Lemma 3 of \cite{bendavid2010}).
\begin{align}
    &|\varepsilon_{\mathcal{D}_{A}}(h,h') - \varepsilon_{\mathcal{D}_{B}}(h,h')| 
    \leq {1 \over 2} d_{\mathcal{H}\Delta\mathcal{H}}(\mathcal{D}_{A}, \mathcal{D}_{B}) \nonumber
\end{align}
Finally, by plugging the Lemma \ref{lem:div_bound} into the above bound, we have Theorem \ref{thm:bound}.

\end{proof}

From Theorem \ref{thm:bound}, we see that the difference between true risks across language domains is bounded by an empirical estimation of the divergence ($d_{\mathcal{H}\Delta\mathcal{H}}$) between those two domains plus an irreducible term defined by problem setup. Thus, if we reduce the divergence between language distributions, the expected performance gap can also be reduced accordingly. To investigate whether this is indeed a case or not, we provided the embedding-space similarity and logit-space Sinkhorn distance between different languages in Figure \ref{fig:qualitative analysis}. We argue that phonemic representation's relatively mild performance gap is achieved by reducing linguistic gaps in the embedding space (high CKA) and final output space (low Sinkhorn distance) those are the proxy of $\mathcal{H}$-divergence.